\documentclass{article} 
\usepackage{iclr2027_conference,times}
\iclrfinalcopy

\usepackage{amsmath,amsfonts,bm}

\def\eqref#1{equation~\ref{#1}}

\def\1{\bm{1}}

\DeclareMathAlphabet{\mathsfit}{\encodingdefault}{\sfdefault}{m}{sl}
\SetMathAlphabet{\mathsfit}{bold}{\encodingdefault}{\sfdefault}{bx}{n}

\usepackage{hyperref}
\usepackage{url}
\usepackage{graphicx}

\usepackage{enumitem}

\usepackage{booktabs}

\title{BridgeRouter: Gain-Aware Instance Routing over Cross-Modal Bridge Portfolios}

\author{
Ruoxi Piao$^{1,*}$,
Alicia Lu$^{1,*}$,
Yunzhen Zhang$^{1,*}$,
Mustafa Mısır$^{1,\dagger}$,
Hasan Onur Keleş$^{2,\dagger}$ \\
$^1$Duke Kunshan University \\
$^2$Ankara University \\
$^*$Equal contribution
\qquad
$^\dagger$Corresponding authors
}

\begin{document}

\maketitle

\begin{abstract}
Lightweight representation bridges make pretrained, task-adapted EEG foundation models executable on other biosignal modalities without retraining their backbones. However, selecting a single bridge pathway globally assigns every instance to the same transfer route and ignores complementary errors across routes. We therefore formulate multi-bridge transfer as an instance-level algorithm-selection problem and introduce \textsc{BridgeRouter}, a gain-aware router over a portfolio of frozen cross-modal pathways. A validation-selected Single Best Solver (SBS) serves as a protected default route. During training, task labels are used to compute the reduction in negative log-likelihood provided by each alternative route relative to the SBS, and a lightweight MLP predicts these route-specific gains from a target-modality representation extracted by a frozen encoder. At inference, \textsc{BridgeRouter} observes only the target modality, requires neither EEG, task labels, nor predictions from all candidate routes, and executes exactly one bridge pathway: the route with the largest predicted gain when that gain exceeds a validation-calibrated threshold, and the SBS otherwise. We evaluate seven EEG foundation-model bridge pathways for EMG-based freezing-of-gait detection on FOG and ECG-based sleep classification on ISRUC. Across five seeds, \textsc{BridgeRouter} achieves $81.38\pm1.11\%$ balanced accuracy on FOG, improving over the BIOT SBS by $11.49$ percentage points, and $62.92\pm2.17\%$ on ISRUC, improving over the CBraMod SBS by $3.83$ points. Per-instance bridge oracles reach $99.96\%$ and $78.61\%$, respectively, revealing substantial remaining portfolio headroom. These results show that anchor-relative gain prediction can convert complementary cross-modal bridge behavior into deployable instance-level routing improvements while retaining single-route inference.
\end{abstract}

\section{Introduction}

What if the best model for transfer is not the best model for every instance?

Cross-modal transfer provides a natural way to reuse knowledge from pretrained biosignal foundation models when the source modality is unavailable at deployment. For example, a pretrained EEG model can be connected to EMG or ECG through a lightweight representation bridge learned from paired recordings~\citep{li2026bioxbridge}. When multiple pretrained EEG models are available, a natural strategy is to build a bridge for each model and simply choose the one with the best average performance.

However, average performance can hide an important property of a transfer portfolio: \emph{different pathways can fail on different instances}. A pathway that is weaker overall may still correctly predict examples that a stronger pathway misses. In such a case, its value does not come from being the best standalone model, but from the errors it can correct. This suggests a different question:

\begin{quote}
\emph{Given the prediction we already have, can another transfer pathway do better on this instance?}
\end{quote}

Answering this question is challenging because the information needed to determine the correct pathway is unavailable at inference. During training, paired EEG–target recordings and task labels reveal which pathways succeed or fail on each instance. At deployment, however, only the target modality is observed. A useful router must therefore learn pathway complementarity from training-time transfer outcomes while making its decisions entirely from target-modality information.

We introduce \textbf{BridgeRouter}, a gain-aware framework for instance-level selection among complementary cross-modal pathways. We construct a portfolio of target-side executable pathways, each combining a frozen pretrained EEG foundation model with a lightweight bridge. One pathway is selected on the validation set as the Single Best Solver and serves as the default route. During training, the router learns the task-label-derived reduction in negative log-likelihood provided by each alternative pathway relative to this anchor. At inference, the router predicts these anchor-relative gains using only the target-modality representation. It selects the most promising alternative when its predicted gain exceeds a validation-calibrated protection threshold and selects the anchor otherwise. The routing decision is made before pathway execution, so exactly one selected bridge pathway is executed for each instance. BridgeRouter therefore treats pathway selection as a conditional decision rather than a global model-selection problem.

We evaluate BridgeRouter on two cross-modal biosignal transfer tasks: adapting pretrained EEG foundation models to EMG inputs for freezing-of-gait detection and to ECG inputs for sleep classification. We examine whether different bridge pathways exhibit complementary errors, whether their anchor-relative gains can be predicted without access to EEG at inference, and whether adaptive routing improves over the validation-selected SBS. Fixed-route performance, per-instance bridge oracles, and rescue-harm diagnostics are used to characterize the available portfolio headroom and the behavior of the learned routing policy.

Our contributions are:
\begin{itemize}[leftmargin=1em] 
    \item We formulate cross-modal transfer with multiple pretrained source models as an instance-level pathway selection problem, emphasizing the complementary errors that distinguish portfolio value from standalone performance.
    
    \item We propose \textbf{BridgeRouter}, which learns task-label-derived, anchor-relative negative-log-likelihood gains from target-modality representations and performs target-modality-only, single-route selection at
    inference.
    
    \item We evaluate gain-aware routing across EEG-to-EMG and EEG-to-ECG transfer tasks and quantify bridge complementarity through fixed-route performance, per-instance oracle headroom, and rescue-harm analyses.
\end{itemize}

\section{Related Work}

\subsection{EEG Foundation Models and Cross-Modal Representation Transfer}
Recent EEG foundation models (FMs) leverage self-supervised pretraining on large-scale unlabeled EEG recordings to learn transferable representations for downstream neural decoding tasks. Representative models, including BENDR~\cite{kostas2021bendr}, BIOT~\cite{yang2023biot}, EEGPT~\cite{wang2024eegpt}, LaBraM~\cite{jiang2024labram}, and CBraMod~\cite{wang2025cbramod}, adopt different architectures, pretraining objectives, and data sources. Recent benchmark studies have further shown substantial variation in their downstream performance across datasets and tasks, suggesting that the choice of a pretrained EEG representation can depend strongly on the target setting~\cite{xiong2026eegfmbench}.

Beyond within-modality adaptation, recent work has explored transferring pretrained biosignal representations across modalities. Lightweight representation bridges provide a practical way to connect a frozen source foundation model to a target biosignal modality without retraining the source model. Such approaches make pretrained representations reusable when the target modality has limited labeled data or when a model pretrained on another modality provides useful prior knowledge. However, when multiple source foundation models are available, each can induce a different cross-modal transfer pathway through its corresponding bridge. Existing cross-modal transfer methods typically evaluate these pathways independently and select a fixed model or bridge based on aggregate performance. They therefore do not explicitly consider whether different pathways make complementary predictions on individual target instances.
\subsection{Model Routing and Per-Instance Model Selection}
A complementary line of work studies conditional model selection, where different inputs can be processed by different models or computational pathways. Mixture-of-experts methods dynamically route inputs to specialized experts, while model cascades and conditional computation adapt the amount or type of computation according to the input~\cite{shazeer2017moe,wang2018skipnet}. More generally, per-instance model selection aims to exploit heterogeneity in model behavior rather than relying on a single model for all inputs.

These approaches motivate adaptive selection in a transfer portfolio, but cross-modal transfer introduces an additional constraint. At training time, paired source–target recordings and task labels can reveal how different pathways behave on the same instance. At deployment, however, the source modality may be unavailable, and executing every source pathway may defeat the purpose of lightweight transfer. Thus, a routing policy must infer the usefulness of alternative pathways from information available on the target side alone. BridgeRouter addresses this setting by predicting anchor-relative pathway gains solely from the target-side representation and selecting a single route before bridge execution, without requiring access to the source modality or executing the entire pathway portfolio at inference.

\subsection{Algorithm Selection and Complementary Model Portfolios}

Automated Algorithm Selection (AS) studies how to select an algorithm from a portfolio according to the characteristics of an individual problem instance~\cite{rice1976algorithm,smithmiles2009meta}. AS has been applied to domains including Boolean satisfiability, combinatorial optimization, and machine learning. A central motivation is that algorithms with different strengths can perform differently across instances, making instance-specific selection potentially more effective than committing to a single globally strongest algorithm. Traditional AS methods often learn performance predictors from handcrafted instance features, while more recent approaches employ learned representations, including convolutional and graph-based encoders, to automatically characterize instances~\cite{loreggia2016deep,cao2025mcGNNASDock}.

Recent work has extended this perspective to foundation-model selection. EEG-AS~\cite{zhang2026eegas} formulates EEG foundation model selection as an instance-level AS problem and learns to estimate foundation-model behavior from inference-available EEG representations and an anchor model. Its formulation demonstrates that foundation-model portfolios can benefit from instance-dependent selection rather than relying exclusively on a Single Best Solver. However, EEG-AS considers selection among pretrained EEG models within the same modality. In contrast, BridgeRouter considers a portfolio of \emph{cross-modal transfer pathways}, where each candidate pathway combines a pretrained source model with a target-side bridge.

This distinction also changes the selection objective. Rather than predicting which pathway is globally best or reconstructing the absolute performance of every candidate, BridgeRouter focuses on the \emph{relative benefit of switching} from an anchor pathway to an alternative pathway. This formulation explicitly exploits complementary errors: an alternative pathway is valuable when it succeeds on instances where the anchor fails, even if its overall performance is lower. BridgeRouter therefore connects automated algorithm selection with cross-modal transfer by treating pathway complementarity as a source of instance-level adaptive performance.
\section{Methods}
\subsection{Problem Setup and Notation}

We consider a target biosignal task on modality $m$, such as EMG or ECG, where knowledge from pretrained EEG foundation models is available. During training, paired samples $(x^e,x^m,y)$ are given, with $x^e$ and $x^m$ denoting synchronized EEG and target-modality observations and $y$ the task label. This paired setting provides the information needed to establish cross-modal transfer. At inference, however, EEG is unavailable, so the final prediction must be obtained from $x^m$ alone.

Our goal is not to transfer one EEG model into the target modality and use it as a fixed predictor. Instead, we construct multiple transfer pathways, each based on a different EEG foundation model and its corresponding cross-modal bridge. Let $F_k^e$ denote the $k$-th frozen EEG foundation model and let
$E^m$ denote the frozen pretrained target-modality encoder. The target-modality
representation available to the router is denoted by
\[
\phi(x^m) = E^m(x^m).
\]
A lightweight bridge $B_k$ connects a representation produced by the target-modality encoder to the representation space of $F_k^e$, defining a pathway $\pi_k$. The resulting pathways form a portfolio $\{\pi_k\}_{k=1}^{K}$, with pathway output $\hat{y}_{ik}$ and class probabilities $p_{ik}$ for instance $i$.

We designate one member of the portfolio as the anchor pathway $\pi_a$. The anchor is the Single Best Solver (SBS), selected exclusively according to validation balanced accuracy:
\[
a
=
\arg\max_{k\in\{1,\ldots,K\}}
\operatorname{BAcc}_{D_{\mathrm{val}}}(\pi_k).
\]
It serves as the default route in the routing decision and is frozen before test evaluation. 

This portfolio introduces a different question from conventional cross-modal transfer. Rather than asking which EEG foundation model transfers best on average, we ask whether the best pathway can vary from one target-modality instance to another. This distinction is important because the router must make its decision without access to the privileged EEG observation. Given only the target representation $\phi(x^m)$, the router $G_\theta$ estimates whether switching from the anchor to another pathway is beneficial. Thus, the routing decision is learned from training-time transfer outcomes but, at inference, depends only on information available from $x^m$.

\subsection{Cross-modal Pathways via Representation Bridging}

The key to constructing such a portfolio is to make each EEG foundation model executable from the target modality alone. We therefore learn a lightweight bridge that maps target-modality representations into the representation space of each frozen EEG foundation model. Following the representation-bridging formulation of BioX-Bridge~\citep{li2026bioxbridge}, the bridge is learned from paired EEG-target observations, while the foundation model itself remains frozen.

For the $k$-th pathway, the target-modality representation is mapped as
\begin{equation}
z_{ik}^e = B_k\!\left(\phi_k(x_i^m)\right),
\end{equation}

where $\phi_k(x_i^m)$ denotes the representation extracted from the selected source layer of the frozen target-modality encoder and $z_{ik}^e$ lies in the representation space of $F_k^e$. During training, the representation extracted from the corresponding EEG observation provides the reference for cross-modal alignment. This alignment stage uses the paired EEG observation only as privileged training information.

Once the bridge is learned, the pathway can be executed without observing EEG. The bridged representation is passed directly to the remaining layers of the frozen EEG foundation model:
\begin{equation}
p_{ik}
=
\operatorname{softmax}
\left[
\operatorname{Head}_k
\left(
F^{e}_{k,>d_k}
\left(
B_k(\phi_k(x_i^m))
\right)
\right)
\right],
\qquad
\hat{y}_{ik}=\arg\max_c p_{ik,c}.
\end{equation}

The resulting computation is therefore driven by $x^m$ even though the frozen backbone originates from EEG. Different choices of $F_k^e$ and $B_k$ consequently give different target-side executable pathways, allowing the pretrained models to contribute complementary transfer behavior rather than being reduced to a single transferred predictor.

\subsection{Portfolio View of Multi-Bridge Transfer}

Because the pathways are built from different foundation models and bridges, they need not make the same prediction on a given target-modality sample. A pathway may capture a particular pattern well while another pathway fails on the same instance. Consequently, their average accuracies do not fully characterize the value of the portfolio: even a weaker pathway on average may provide the correct prediction on samples where the anchor fails.

We therefore evaluate transfer quality at the instance level. For a sample $i$, let
\begin{equation}
\label{eq:correctness}
c_{ik} = \mathbf{1}\!\left[\hat y_{ik}=y_i\right].
\end{equation}
This variable indicates whether pathway $k$ predicts the label correctly. The corresponding portfolio oracle is
\begin{equation}
c_i^{*} = \max_{k} c_{ik},
\end{equation}
which indicates whether at least one available pathway can solve the instance correctly.

The gap between a fixed pathway and this oracle exposes the opportunity for routing: the portfolio contains useful predictions that a single globally selected pathway cannot recover. The problem is therefore not to determine one universally preferred bridge, but to identify, for each target-modality instance, whether another pathway is more suitable than the anchor. BridgeRouter addresses this problem by learning this instance-level pathway preference from training-time transfer outcomes, while keeping the routing decision entirely target-modality based at inference.

\subsection{BridgeRouter: Training and Inference}

\begin{figure}[htb!]
    \centering
    \includegraphics[width=\linewidth]{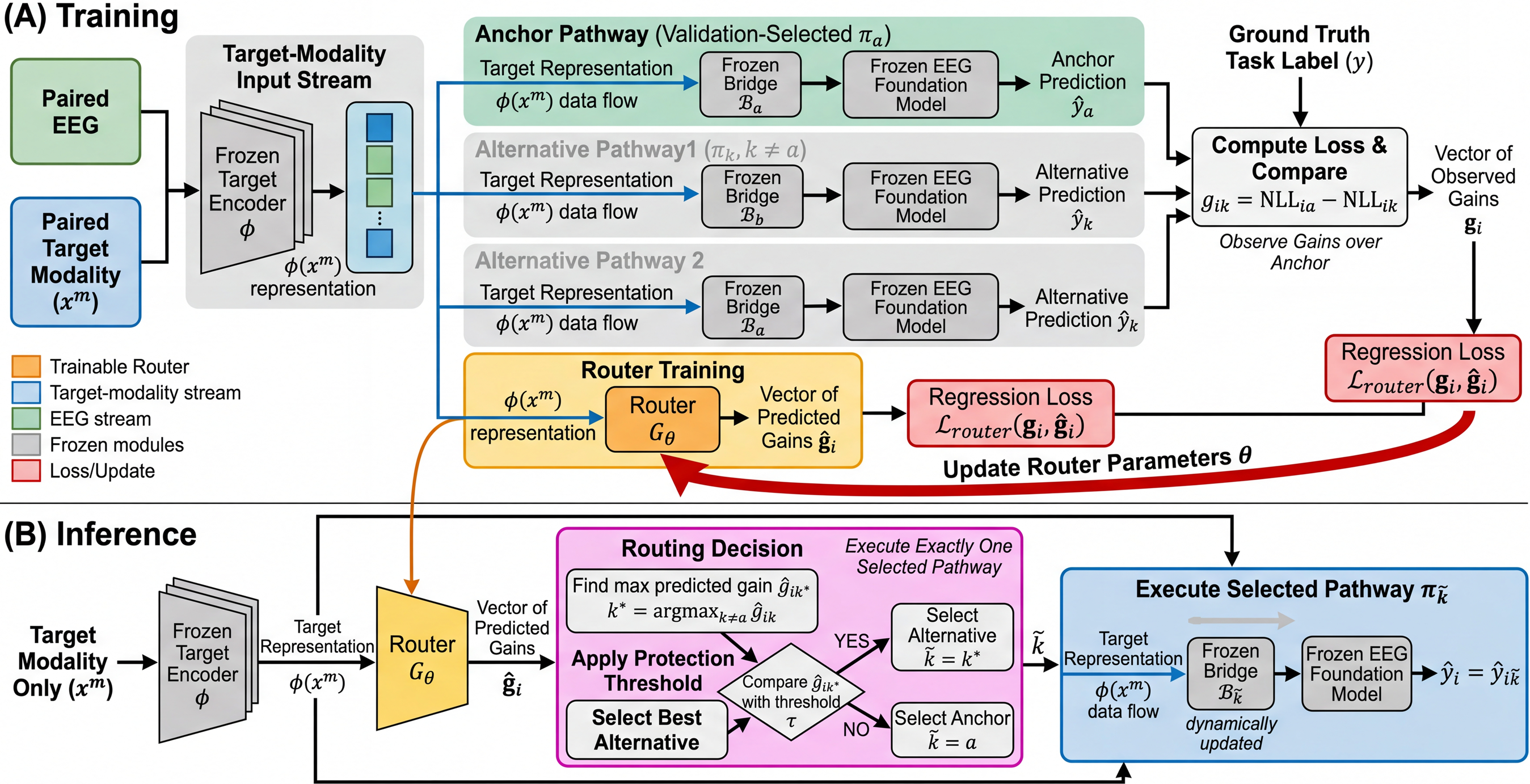}
\caption{
Overview of BridgeRouter.
\textbf{(A) Training.}
Paired EEG--target-modality data are used to train cross-modal bridges and the router, with the validation-selected pathway $\pi_a$ serving as the anchor.
The router is supervised by the anchor-relative gain
$g_{ik}=\mathrm{NLL}_{ia}-\mathrm{NLL}_{ik}$.
\textbf{(B) Inference.}
Given only the target modality, the router predicts the gains of alternative pathways and selects the best alternative when its predicted gain exceeds the protection threshold $\tau$; otherwise, it retains the anchor.
}
    \label{fig:framework}
\end{figure}

Figure~\ref{fig:framework} summarizes the proposed BridgeRouter framework. 
As discussed above, different pathways can be preferable for different target-modality instances, making a fixed choice of pathway insufficient. 
At inference time, however, the ground-truth label is unavailable, so we cannot directly determine which pathway will provide the largest improvement. 
This limitation raises a practical question: given a particular target-modality instance, which pathway should we use?
BridgeRouter addresses this question by predicting the benefit of each alternative route from the target-modality representation and selecting either the anchor or one alternative before executing a bridge pathway.

\paragraph{Training.}
The key question is not whether an alternative pathway is accurate in isolation, but whether it can improve upon the pathway we would use by default. 
We therefore train the router to predict each alternative pathway's loss reduction relative to the anchor. During router training, all pathways can be evaluated on the training data, allowing us to measure how much each alternative improves upon the anchor for each instance.

For sample $i$, let \begin{equation}
\operatorname{NLL}_{ik}
=
-\log p_{ik,y_i}
\end{equation}
denote the negative log-likelihood of the prediction produced by pathway $k$. 
Using the anchor as the reference, we define the observed benefit of switching to an alternative pathway $k\neq a$ as
\begin{equation}
g_{ik}
=
\operatorname{NLL}_{ia}
-
\operatorname{NLL}_{ik}.
\end{equation}
A positive $g_{ik}$ indicates that pathway $k$ improves upon the anchor prediction for that instance, while a negative value indicates that switching to pathway $k$ increases the prediction loss.

The router learns to predict the switching benefits from information available at deployment,
\begin{equation}
\hat{\mathbf g}_i
=
G_\theta\!\left(\phi(x_i^m)\right),
\end{equation}
where $\hat{\mathbf g}_i$ contains the predicted gains for all alternative pathways $k\neq a$. 
We train the router using a regression loss between $\hat{\mathbf g}_i$ and the observed gains $\mathbf g_i$. 
The paired EEG observation is used to establish the cross-modal bridges, while the task label is used to compute the route-specific negative log-likelihood gains. Neither the EEG observation nor the task label is provided as an input to the router. At inference, the router requires only the target-modality representation.

\paragraph{Inference.}
At deployment, the router computes the target-modality representation $\phi(x_i^m)$ and predicts the relative gain of each alternative pathway:
\begin{equation}
\hat{\mathbf g}_i
=
G_\theta\!\left(\phi(x_i^m)\right).
\end{equation}
The router then identifies the most promising alternative:
\begin{equation}
k_i^*
=
\arg\max_{k\neq a}\hat g_{ik}.
\end{equation}

A large predicted gain alone, however, does not necessarily justify switching away from the validation-selected SBS. We therefore use a protection threshold $\tau$, selected exclusively on the validation split, to distinguish a potentially useful correction from a marginal predicted improvement. The
selected route is
\begin{equation}
\tilde{k}_i
=
\begin{cases}
k_i^*, & \hat g_{i k_i^*}>\tau,\\
a, & \hat g_{i k_i^*}\leq\tau.
\end{cases}
\end{equation}

After this routing decision, exactly one pathway, $\pi_{\tilde{k}_i}$, is executed using the target-modality input. The final
prediction is
\begin{equation}
\hat y_i
=
\hat y_{i\tilde{k}_i}.
\end{equation}

Thus, the anchor serves as the default route: BridgeRouter switches to an alternative only when its predicted gain exceeds the validation-selected threshold. The anchor does not need to be executed before routing because the router conditions only on the target-modality representation. BridgeRouter therefore retains SBS protection while requiring exactly one bridge-pathway execution per instance.

\section{Experiments and Discussion}
\subsection{Experimental Setup}

\paragraph{Datasets.}
We evaluate on two multimodal physiological-signal datasets:
the freezing-of-gait (\textbf{FOG}) dataset~\cite{zhang2022fog} and
the \textbf{ISRUC} sleep dataset~\cite{khalighi2016isruc}.
FOG contains synchronized EEG and EMG recordings from 12 participants.
We use three-second windows and obtain 37,200 paired EEG--EMG windows
after preprocessing for binary FOG classification.
For ISRUC, we use Subgroup I with synchronized EEG and ECG recordings.
Following BioX-Bridge~\cite{li2026bioxbridge}, the original five sleep
stages are grouped into Wake versus Sleep, with each 30-second epoch
treated as one sample. We use six EEG derivations
(F3--A2, C3--A2, O1--A2, F4--A1, C4--A1, and O2--A1).
Thus, EEG is the old modality, while EMG and ECG are the new modalities
for FOG and ISRUC, respectively. Further preprocessing details are given
in Appendix~\ref{app:experimental-setup}.

\paragraph{Data Partitioning.}
We follow the partitioning protocol of BioX-Bridge~\cite{li2026bioxbridge}.
Each dataset is divided into $D_{\mathrm{old}}$, $D_{\mathrm{pair}}$,
$D_{\mathrm{val}}$, and $D_{\mathrm{new}}$, with approximate proportions
of 33\%, 33\%, 11\%, and 22\%. $D_{\mathrm{old}}$ is used to train
task-specific heads for the frozen EEG models, $D_{\mathrm{pair}}$ is used
for bridge and routing training, $D_{\mathrm{val}}$ is used for model
selection, and $D_{\mathrm{new}}$ is used only for final evaluation.
ISRUC is split by subject, so no subject appears in more than one subset.
FOG follows the sample-wise BioX protocol and therefore does not evaluate
unseen-subject generalization. The detailed FOG split is reported in
Appendix~\ref{app:experimental-setup}.

\paragraph{Foundation Models and Bridge Routes.}
We evaluate seven pretrained EEG foundation models:
LaBraM~\cite{jiang2024labram}, EEGPT~\cite{wang2024eegpt},
REVE~\cite{el2026reve}, BIOT~\cite{yang2023biot},
CBraMod~\cite{wang2025cbramod}, CSBrain~\cite{zhou2026csbrain},
and BENDR~\cite{kostas2021bendr}.
We freeze all EEG backbones and train a linear task head on
$D_{\mathrm{old}}$. For the new modalities, we use pretrained NormWear
for EMG~\cite{luo2025normwear} and HuBERT-ECG for ECG~\cite{coppola2024hubertecg},
which are also frozen. Each EEG model defines one cross-modal bridge,
giving seven candidate routes per dataset. Bridge layer positions are
selected following BioX-Bridge.

\paragraph{Baselines and Comparators.}
We first evaluate each bridge pathway independently. The single best solver (SBS) is selected exclusively by validation balanced accuracy and frozen before test evaluation. This procedure selects BIOT on FOG and CBraMod on ISRUC. The SBS is never reselected according to test performance.

Our primary method is \textsc{BridgeRouter}, a gain-aware instance-level routing framework over cross-modal bridge pathways. It predicts the instance-specific reduction in negative log-likelihood of each candidate route relative to the validation-selected SBS. The router retains the SBS unless the largest predicted gain exceeds a threshold selected using validation data. It uses task-label-derived gain targets during training but requires neither labels nor bridge predictions at inference. It observes only the target-modality representation and executes one selected bridge pathway.

We additionally report fixed-route performance and per-instance bridge oracles. The fixed routes establish the performance of each available pathway without routing. The oracles quantify portfolio complementarity by selecting a correct pathway using the unknown ground-truth label. They are non-deployable and are not used for training, model selection, or calibration.

\paragraph{Implementation Details.}
\textsc{BridgeRouter} uses the pooled NormWear block-6 representation on FOG and the 512-dimensional HuBERT-ECG block-4 representation on ISRUC. Its MLP contains hidden dimensions 256 and 128, GELU activations, and dropout of 0.2. The model predicts the route-specific negative log-likelihood gain relative to the validation-selected SBS and is trained using class-balanced supervision.

For each instance, the route with the largest predicted gain is identified. The SBS is retained unless this gain exceeds a protection threshold selected on the validation split. At inference, the router observes only the target-modality representation and executes one bridge pathway. It does not require the test label or the predictions of all candidate bridges. We evaluate seeds 42-46 while keeping all bridge checkpoints and data partitions fixed.

\paragraph{Evaluation Metrics.}
Balanced accuracy is the primary metric because both tasks are class-imbalanced. We additionally report accuracy and macro F1. For routing diagnostics, a rescue occurs when the SBS is incorrect and the selected alternative is correct; a harm occurs when the SBS is correct and the selected alternative is incorrect. All stochastic routing results report mean and sample standard deviation over five seeds.

\subsection{Frozen Bridge Performance}

Table~\ref{tab:bridge_performance} reports the performance of every frozen cross-modal pathway on the two final test sets. Bridge quality varies substantially across receivers and does not directly follow native EEG performance.

On FOG, BIOT is both the validation-selected SBS and the strongest test pathway, reaching 69.89\% balanced accuracy. BENDR and CSBrain follow at 65.96\% and 64.05\%, whereas LaBraM, EEGPT, and REVE exhibit near-chance and strongly class-biased bridge predictions.

On ISRUC, CBraMod is selected as the SBS because it obtains the highest validation balanced accuracy. Its test balanced accuracy, however, drops to 59.09\%. EEGPT and BENDR reach 65.17\% and 64.45\% on test, respectively. These results do not change the SBS retrospectively: selecting EEGPT after observing test labels would constitute test-set model selection. Instead, the ranking reversal demonstrates that a globally selected pathway may fail to generalize across unseen subjects.




\begin{table}[htb!]
\caption{Frozen cross-modal bridge performance on the final test sets.
$\dagger$ denotes the validation-selected SBS. All values are percentages.}
\label{tab:bridge_performance}
\centering
\small
\setlength{\tabcolsep}{5pt}

\begin{minipage}[t]{0.48\linewidth}
\centering
\textbf{(a) FOG}\par\vspace{2pt}
\begin{tabular}{l | rrr}
\toprule
Receiver & Accuracy & BAcc & Macro-F1 \\
\midrule
LaBraM  & 57.63 & 49.98 & 36.59 \\
EEGPT   & 58.39 & 51.09 & 39.83 \\
REVE    & 42.91 & 50.41 & 31.14 \\
BIOT$^\dagger$
        & \textbf{69.07} & \textbf{69.89} & \textbf{68.98} \\
CBraMod & 57.30 & 58.02 & 57.23 \\
CSBrain & 65.07 & 64.05 & 64.10 \\
BENDR   & 65.72 & 65.96 & 65.47 \\
\bottomrule
\end{tabular}
\end{minipage}
\hfill
\begin{minipage}[t]{0.48\linewidth}
\centering
\textbf{(b) ISRUC}\par\vspace{2pt}
\begin{tabular}{l | rrr}
\toprule
Receiver & Accuracy & BAcc & Macro-F1 \\
\midrule
LaBraM  & 78.087 & 60.486 & 61.772 \\
CBraMod$^\dagger$
        & 78.087 & 59.090 & 60.229 \\
REVE    & 79.693 & 61.732 & 63.445 \\
EEGPT   & 75.920 & \textbf{65.171} & 65.000 \\
CSBrain & 79.096 & 62.905 & 64.438 \\
BENDR   & \textbf{81.506} & 64.445 & \textbf{66.755} \\
BIOT    & 80.518 & 64.051 & 66.005 \\
\bottomrule
\end{tabular}
\end{minipage}

\end{table}

The pathway errors remain complementary on both datasets. The seven-route oracle reaches 99.96\% balanced accuracy on FOG and 78.61\% on ISRUC, leaving 30.07 and 19.52 percentage points of headroom over the respective validation-selected SBSs. The FOG oracle is partly inflated by complementary class-biased routes, but the validation-filtered compact-four oracle still reaches 93.69\%. Therefore, substantial complementarity remains even after removing collapsed pathways.

\begin{table}[htb!]
\caption{Bridge-portfolio headroom on the final test sets. Oracles are
non-deployable.}
\label{tab:portfolio_headroom}
\centering
\small
\begin{tabular}{l | lrrr}
\toprule
Dataset & SBS & SBS BAcc & K7 Oracle BAcc & Headroom \\
\midrule
FOG   & BIOT    & 69.89 & 99.96 & $+30.07$ \\
ISRUC & CBraMod & 59.09 & 78.61 & $+19.52$ \\
\bottomrule
\end{tabular}
\end{table}

\subsection{Gain-Aware Bridge Routing}

Table~\ref{tab:gain_routing} presents the primary cross-dataset result. On FOG, the \textsc{BridgeRouter} reaches $81.38\pm1.11\%$ balanced accuracy, improving over the BIOT SBS by $11.49$ percentage points. On ISRUC, it reaches $62.92\pm2.17\%$, improving over the validation-selected CBraMod SBS by $3.83$ points. Every ISRUC seed improves over CBraMod in balanced accuracy.

The ISRUC improvement is smaller and more variable than the FOG improvement. This is consistent with its subject-disjoint protocol and the pronounced validation-to-test pathway-ranking shift. The result shows that relative-gain supervision transfers across the two tasks, but it does not imply dominance over every fixed test pathway. In particular, EEGPT and BENDR obtain higher ISRUC test balanced accuracy than the five-seed router mean, although neither was selected as the SBS using validation balanced accuracy.

\begin{table}[t]
\caption{Primary \textsc{BridgeRouter} results on the final test
sets. Router values are mean $\pm$ sample standard deviation over five
seeds. $\Delta$ is measured relative to the validation-selected SBS.
Oracles are non-deployable.}
\label{tab:gain_routing}
\centering
\small
\setlength{\tabcolsep}{4.5pt}
\begin{tabular}{l | l | rrrr}
\toprule
Dataset & Method & Accuracy & BAcc & Macro-F1 & $\Delta$BAcc \\
\midrule
FOG
& BIOT SBS
& 69.07 & 69.89 & 68.98 & -- \\

FOG
& \textsc{BridgeRouter}
& $\mathbf{81.49\pm1.20}$
& $\mathbf{81.38\pm1.11}$
& $\mathbf{81.17\pm1.18}$
& $\mathbf{+11.49}$ \\

FOG
& K7 Oracle
& 99.96 & 99.96 & -- & $+30.07$ \\
\hline

ISRUC
& CBraMod SBS
& 78.09 & 59.09 & 60.23 & -- \\

ISRUC
& \textsc{BridgeRouter}
& $\mathbf{78.66\pm0.75}$
& $\mathbf{62.92\pm2.17}$
& $\mathbf{64.25\pm2.25}$
& $\mathbf{+3.83}$ \\

ISRUC
& K7 Oracle
& 90.13 & 78.61 & 82.98 & $+19.52$ \\
\bottomrule
\end{tabular}
\end{table}

Relative to the corresponding SBS-to-oracle gaps, the router recovers approximately 38.2\% of the available FOG headroom and 19.6\% of the ISRUC headroom. At inference, the router observes only the target-modality representation, requires no test label, and executes one bridge pathway. Thus, its computational cost is substantially lower than evaluating all seven routes.

The ISRUC result must nevertheless be interpreted conservatively. Its five-seed switch rate is highly variable, largely because one seed switches substantially more often than the remaining four. Moreover, the router mean remains below the strongest fixed test pathways. The consistent improvement over the validation-selected SBS demonstrates useful instance-level gain prediction, while the remaining oracle gap shows that reliable cross-subject route selection remains unresolved.

\subsection{Cross-Dataset Findings}

The two datasets support three common conclusions.

First, frozen bridge pathways exhibit substantial but heterogeneous complementarity. Neither native EEG performance nor a single global bridge ranking fully describes instance-level pathway behavior. The seven-route oracle reaches 99.96\% balanced accuracy on FOG and 78.61\% on ISRUC, leaving substantial headroom over the validation-selected SBS on both datasets.

Second, predicting relative task loss provides useful routing supervision across both clinical tasks. The \textsc{BridgeRouter} improves over the validation-selected SBS by 11.49 balanced-accuracy points on FOG and by 3.83 points on ISRUC. Because only one selected bridge is executed, this improvement does not require evaluating the complete portfolio at inference.

Third, protocol differences materially affect routing difficulty. FOG uses a sample-wise split containing the same participants across subsets, whereas ISRUC evaluates unseen subjects. ISRUC also exhibits a substantial validation-to-test pathway-ranking reversal: CBraMod is the validation-selected SBS but is not the strongest individual test pathway. Consequently, improvement over the correctly selected SBS can coexist with lower performance than pathways identified only after observing test results.

These findings establish \textsc{BridgeRouter} as the primary cross-dataset method while showing that robust instance-level route selection under cross-subject distribution shift remains an open problem.

\section{Conclusion}

We presented \textsc{BridgeRouter}, an instance-level routing framework for cross-modal transfer from pretrained EEG foundation models. Rather than relying on a single globally selected bridge, our approach learns to predict the relative task-loss gain of candidate bridges and selects an alternative route only when the predicted gain is sufficiently large. This enables instance-adaptive bridge selection while executing only one bridge pathway at inference time.

Across the FOG and ISRUC datasets, we found substantial complementarity among frozen EEG-to-target-modality bridges. The proposed router improves over the validation-selected single best solver by 11.49 and 3.83 percentage points in balanced accuracy on FOG and ISRUC, respectively. At the same time, the remaining gap to the oracle and the smaller and more variable routing gains under the subject-disjoint ISRUC setting highlight the difficulty of reliable instance-level routing under distribution shift.

These results suggest that cross-modal transfer can benefit from treating pretrained foundation models as a portfolio of complementary bridge routes rather than committing to a single global choice. Future work will investigate more robust routing under cross-subject and cross-domain shifts, as well as richer signals for estimating when a bridge is likely to provide a meaningful transfer gain.

\subsection*{AI Use Statement}

Generative AI tools were used to assist with language editing and polishing of the manuscript, including improving clarity, grammar, and organization. They were not used to generate research hypotheses, determine the experimental methodology, conduct experiments, or make scientific conclusions. All AI-assisted text was reviewed and revised by the authors, who take full responsibility for the content of the manuscript.

\subsection*{Ethics Statement}

This work uses existing physiological datasets for methodological research on EEG foundation models. We did not collect new data or interact directly with human participants. All experiments were conducted using the datasets under their respective access and usage conditions. The study focuses on methodological evaluation and does not make clinical decisions or provide medical recommendations. We report both positive and negative experimental findings and describe the evaluation protocol to reduce the risk of misleading conclusions arising from model selection or test-set reuse.

\subsection*{Reproducibility Statement}

We describe the main experimental settings, model configurations, and evaluation procedures used in this study in the main paper and appendix. Where applicable, we report the foundation-model checkpoints, bridge configurations, and random seeds used in our experiments. We also provide implementation details necessary to reproduce the proposed method and the reported evaluation protocol.

\bibliography{iclr2027_conference}

@article{zhang2022fog,
  title   = {Multimodal Data for the Detection of Freezing of Gait in Parkinson's Disease},
  author  = {Zhang, W. and Yang, Z. and Li, H. and Huang, D. and Wang, L. and Wei, Y. and Zhang, L. and Ma, L. and Feng, H. and Pan, J. and Guo, Y. and Chan, P.},
  journal = {Scientific Data},
  volume  = {9},
  number  = {1},
  pages   = {606},
  year    = {2022},
  doi     = {10.1038/s41597-022-01713-8}
}

@article{khalighi2016isruc,
  title   = {{ISRUC-Sleep}: A Comprehensive Public Dataset for Sleep Researchers},
  author  = {Khalighi, S. and Sousa, T. and Santos, J. M. and Nunes, U.},
  journal = {Computer Methods and Programs in Biomedicine},
  volume  = {124},
  pages   = {180--192},
  year    = {2016},
  doi     = {10.1016/j.cmpb.2015.10.013}
}

@inproceedings{li2026bioxbridge,
  title     = {{BioX-Bridge}: Model Bridging for Unsupervised Cross-Modal Knowledge Transfer across Biosignals},
  author    = {Li, Chenqi and Liu, Yu and Denison, Timothy and Zhu, Tingting},
  booktitle = {International Conference on Learning Representations},
  pages     = {111389--111415},
  year      = {2026},
  url       = {https://proceedings.iclr.cc/paper_files/paper/2026/file/b4ec8a0ae0373d694dc8529ca523e9a9-Paper-Conference.pdf}
}

@inproceedings{xiong2026eegfmbench,
  title     = {{EEG-FM-Bench}: A Comprehensive Benchmark for the Systematic Evaluation and Diagnostic Analyses of {EEG} Foundation Models},
  author    = {Xiong, Wei and Li, Jiangtong and Li, Jie and Zhu, Kun and Jiang, Changjun},
  booktitle = {Forty-third International Conference on Machine Learning},
  year      = {2026},
  url       = {https://openreview.net/forum?id=vGeNaFHdET}
}

@inproceedings{shazeer2017moe,
  title     = {Outrageously Large Neural Networks: The Sparsely-Gated Mixture-of-Experts Layer},
  author    = {Shazeer, Noam and Mirhoseini, Azalia and Maziarz, Krzysztof and Davis, Andy and Le, Quoc V. and Hinton, Geoffrey and Dean, Jeff},
  booktitle = {International Conference on Learning Representations},
  year      = {2017}
}

@inproceedings{wang2018skipnet,
  title     = {{SkipNet}: Learning Dynamic Routing in Convolutional Networks},
  author    = {Wang, Xin and Yu, Fisher and Dou, Zi-Yi and Darrell, Trevor and Gonzalez, Joseph E.},
  booktitle = {European Conference on Computer Vision},
  pages     = {409--424},
  year      = {2018}
}

@article{rice1976algorithm,
  title   = {The Algorithm Selection Problem},
  author  = {Rice, John R.},
  journal = {Advances in Computers},
  volume  = {15},
  pages   = {65--118},
  year    = {1976},
  doi     = {10.1016/S0065-2458(08)60520-3}
}

@article{smithmiles2009meta,
  title   = {Cross-disciplinary Perspectives on Meta-learning for Algorithm Selection},
  author  = {Smith-Miles, Kate},
  journal = {ACM Computing Surveys},
  volume  = {41},
  number  = {1},
  year    = {2009},
  doi     = {10.1145/1456650.1456656}
}

@inproceedings{loreggia2016deep,
  title     = {Deep Learning for Algorithm Portfolios},
  author    = {Loreggia, Andrea and Malitsky, Yuri and Samulowitz, Horst and Saraswat, Vijay},
  booktitle = {Proceedings of the Thirtieth AAAI Conference on Artificial Intelligence},
  volume    = {30},
  number    = {1},
  year      = {2016},
  doi       = {10.1609/aaai.v30i1.10170}
}

@inproceedings{cao2025mcGNNASDock,
  title     = {{MC-GNNAS-Dock}: Multi-criteria GNN-based Algorithm Selection for Molecular Docking},
  author    = {Cao, Siyuan and Wu, Hongxuan and Wang, Jiabao Brad and Yuan, Yiliang and Misir, Mustafa},
  booktitle = {PRICAI 2025: Trends in Artificial Intelligence},
  pages     = {686--696},
  year      = {2025},
  doi       = {10.1007/978-981-95-7075-1_46}
}

@misc{zhang2026eegas,
  title         = {{EEG-AS}: Instance-Level Foundation Model Selection for {EEG} Foundation Models via Behavior Reconstruction},
  author        = {Zhang, Yunzhen and Piao, Ruoxi and Keles, Hasan Onur and Misir, Mustafa},
  year          = {2026},
  eprint        = {2609.00653},
  archivePrefix = {arXiv},
  primaryClass  = {cs.LG},
  url           = {https://arxiv.org/abs/2609.00653}
}

@article{el2026reve,
  title={REVE: A foundation model for EEG-adapting to any setup with large-scale pretraining on 25,000 subjects},
  author={El Ouahidi, Yassine and Lys, Jonathan and Th{\"o}lke, Philipp and Farrugia, Nicolas and Pasdeloup, Bastien and Gripon, Vincent and Jerbi, Karim and Lioi, Giulia},
  journal={Advances in Neural Information Processing Systems (NeurIPS)},
  volume={38},
  pages={22541--22577},
  year={2026}
}

@inproceedings{wang2025cbramod,
  title={Cbramod: A criss-cross brain foundation model for eeg decoding},
  author={Wang, Jiquan and Zhao, Sha and Luo, Zhiling and Zhou, Yangxuan and Jiang, Haiteng and Li, Shijian and Li, Tao and Pan, Gang},
  booktitle={International conference on learning representations},
  volume={2025},
  pages={75310--75346},
  year={2025}
}

@article{yang2023biot,
  title={Biot: Biosignal transformer for cross-data learning in the wild},
  author={Yang, Chaoqi and Westover, M and Sun, Jimeng},
  journal={Advances in Neural Information Processing Systems (NeurIPS)},
  volume={36},
  pages={78240--78260},
  year={2023}
}

@article{zhou2026csbrain,
  title={Csbrain: A cross-scale spatiotemporal brain foundation model for eeg decoding},
  author={Zhou, Yuchen and Wu, Jiamin and Ren, Zichen and Yao, Zhouheng and Lu, Weiheng and Peng, Kunyu and Zheng, Qihao and Song, Chunfeng and Ouyang, Wanli and Gou, Chao},
  journal={Advances in Neural Information Processing Systems (NeurIPS)},
  volume={38},
  pages={87150--87195},
  year={2026}
}

@article{wang2024eegpt,
  author  = {Guangyu Wang and Wenchao Liu and Yuhong He and Cong Xu and Lin Ma and Haifeng Li},
  title   = {EEGPT: Pretrained Transformer for Universal and Reliable Representation of EEG Signals},
  journal = {Advances in Neural Information Processing Systems},
  volume  = {37},
  pages   = {39249--39280},
  year    = {2024}
}

@inproceedings{jiang2024labram,
  author    = {Wei-Bang Jiang and Li-Ming Zhao and Bao-Liang Lu},
  title     = {Large Brain Model for Learning Generic Representations with Tremendous EEG Data in BCI},
  booktitle = {International Conference on Learning Representations (ICLR)},
  year      = {2024}
}

@article{kostas2021bendr,
  author  = {Demetres Kostas and Stephane Aroca-Ouellette and Frank Rudzicz},
  title   = {BENDR: Using Transformers and Large Scale Self-Supervised Learning for Generalizable Neural Decoding},
  journal = {Frontiers in Human Neuroscience},
  volume  = {15},
  pages   = {653659},
  year    = {2021}
}

@misc{luo2025normwear,
  title         = {Toward Foundation Model for Multivariate Wearable Sensing of Physiological Signals},
  author        = {Luo, Yunfei and Chen, Yuliang and Salekin, Asif and Rahman, Tauhidur},
  year          = {2025},
  eprint        = {2412.09758},
  archivePrefix = {arXiv},
  primaryClass  = {cs.LG},
  url           = {https://arxiv.org/abs/2412.09758}
}

@article{coppola2024hubertecg,
  title   = {{HuBERT-ECG}: A Self-Supervised Foundation Model for Broad and Scalable Cardiac Applications},
  author  = {Coppola, Edoardo and Savardi, Mattia and Massussi, Mauro and Adamo, Marianna and Metra, Marco and Signoroni, Alberto},
  journal = {medRxiv},
  year    = {2024}
}
\bibliographystyle{iclr2027_conference}

\appendix
\section{Appendix}
\subsection{Experimental Setup}
\label{app:experimental-setup}
\paragraph{Datasets.}
We evaluate our method on two multimodal physiological-signal benchmarks: the freezing-of-gait (\textbf{FOG}) dataset~\cite{zhang2022fog} and the \textbf{ISRUC} sleep dataset~\cite{khalighi2016isruc}. These datasets represent two distinct clinical prediction settings and provide synchronized recordings from multiple physiological modalities.

The FOG dataset contains synchronized EEG and EMG recordings from 12 participants performing gait-related tasks. We segment each recording into three-second windows. The EEG signals are resampled to 200~Hz, producing six-channel windows with 600 temporal samples. The synchronized EMG windows are processed using the preprocessing pipeline associated with the pretrained NormWear encoder. After preprocessing and quality filtering, the FOG dataset contains 37,200 paired EEG--EMG windows for binary FOG classification.

ISRUC is a polysomnographic sleep-staging dataset containing synchronized EEG and ECG recordings from multiple participants. We use Subgroup I of the ISRUC-Sleep dataset, which contains overnight polysomnographic recordings with synchronized EEG and ECG signals. Sleep stages are annotated at 30-second resolution. We use the six EEG derivations F3--A2, C3--A2, O1--A2, F4--A1, C4--A1, and O2--A1, together with the synchronized ECG signal. Each annotated 30-second epoch is treated as one sample for the downstream binary classification task. Following BioX-Bridge~\cite{li2026bioxbridge}, we collapse the original five sleep stages into a binary Wake-versus-Sleep task, where Wake is retained as one class and all sleep stages are grouped into the other.

In both datasets, we consider EEG as the old modality, and EMG and ECG as the new modalities for FOG and ISRUC, respectively.

\paragraph{Data Partitioning Protocol.}
For direct comparison with BioX-Bridge~\cite{li2026bioxbridge}, we follow its dataset-specific partitioning protocol. For each dataset, the available recordings are divided into four subsets: $D_{\mathrm{old}}$, $D_{\mathrm{pair}}$, $D_{\mathrm{val}}$, and $D_{\mathrm{new}}$, with approximate proportions of 33\%, 33\%, 11\%, and 22\%, respectively. Here, $D_{\mathrm{old}}$ provides labeled observations from the old modality for training the task-specific heads of the frozen foundation models; $D_{\mathrm{pair}}$ contains synchronized old- and new-modality observations for learning the cross-modal bridges and transfer signals; $D_{\mathrm{val}}$ is used for model selection and calibration; and $D_{\mathrm{new}}$ is reserved for final evaluation. For ISRUC, partitioning is performed at the subject level, such that all epochs from a given participant are assigned to a single subset and no participant is shared across $D_{\mathrm{old}}$, $D_{\mathrm{pair}}$, $D_{\mathrm{val}}$, and $D_{\mathrm{new}}$.

The resulting FOG split is summarized in Table~\ref{tab:fog_protocol}. The FOG evaluation follows the BioX-style sample-wise protocol and is therefore not an unseen-subject evaluation.

\begin{table}[t]
\caption{FOG12 data partitioning. All four subsets contain samples from the same 12 participants.}
\label{tab:fog_protocol}
\begin{center}
\small
\begin{tabular}{lrrrr}
\toprule
\multicolumn{1}{c}{\bf SPLIT} &
\multicolumn{1}{c}{\bf SAMPLES} &
\multicolumn{1}{c}{\bf CLASS 0} &
\multicolumn{1}{c}{\bf CLASS 1} &
\multicolumn{1}{c}{\bf POS.}
\\ \midrule
$D_{\mathrm{old}}$  & 12,400 & 7,154 & 5,246 & 42.31\% \\
$D_{\mathrm{pair}}$ & 12,400 & 7,025 & 5,375 & 43.35\% \\
$D_{\mathrm{val}}$  & 4,133  & 2,398 & 1,735 & 41.98\% \\
$D_{\mathrm{new}}$  & 8,267  & 4,766 & 3,501 & 42.35\% \\
\bottomrule
\end{tabular}
\end{center}
\end{table}

\paragraph{EEG Foundation Models and Bridge Routes.}
We evaluate seven EEG foundation models: LaBraM~\cite{jiang2024labram}, EEGPT~\cite{wang2024eegpt}, REVE~\cite{el2026reve}, BIOT~\cite{yang2023biot}, CBraMod~\cite{wang2025cbramod}, CSBrain~\cite{zhou2026csbrain}, and BENDR~\cite{kostas2021bendr}. Each model is initialized using its publicly released pretrained weights. For both datasets, we freeze the EEG backbone and train only a task-specific linear classification head on $D_{\mathrm{old}}$. Both FOG and ISRUC are formulated as binary classification tasks.

For the new-modality encoders, we use pretrained NormWear~\cite{luo2025normwear} to encode EMG in FOG and pretrained HuBERT-ECG~\cite{coppola2024hubertecg} to encode ECG in ISRUC. Both encoders remain frozen throughout bridge training and evaluation. We construct one bridge for each EEG foundation model, yielding seven candidate routes per dataset. For every route, we follow the layer-position selection procedure of BioX-Bridge to determine the source layer in the new-modality encoder and the destination layer in the corresponding EEG foundation model.

\subsection{Task-Label-Free Transfer-Aware Routing}
We investigate whether bridge-pathway selection can be learned without task labels or label-derived correctness and gain targets. For each pathway, the task-label-free router predicts three signals from the target-modality representation: native-teacher competence, normalized teacher-to-bridge Jensen-Shannon divergence, and teacher-bridge prediction-flip probability.

The prespecified transfer-aware score is
\[
\hat{s}_{ik}
=
\hat{c}_{ik}(1-\hat{f}_{ik})
-
\hat{d}_{ik},
\]
where $\hat{c}_{ik}$ is predicted teacher competence, $\hat{f}_{ik}$ is predicted flip probability, and $\hat{d}_{ik}$ is predicted normalized Jensen-Shannon divergence. We additionally evaluate competence-only and transfer-only variants to isolate the contribution of the two signal families.

The FOG router uses the validation-filtered compact-four portfolio consisting of BIOT, CBraMod, CSBrain, and BENDR. The ISRUC router uses the complete seven-route portfolio. Router checkpoint selection is based on held-out transfer-signal prediction loss rather than downstream classification performance.

\begin{table}[htb!]
\caption{Task-label-free routing over five seeds. FOG is evaluated on TEST using the compact-four portfolio. ISRUC is evaluated on VAL using the canonical seven-route portfolio. $\Delta$ is measured relative to the dataset-specific validation-selected SBS.}
\label{tab:appendix_label_free_routing}
\centering
\small
\setlength{\tabcolsep}{4.5pt}
\begin{tabular}{l | llrrr}
\toprule
Dataset & Split & Method & Accuracy & BAcc & $\Delta$BAcc \\
\midrule
FOG
& TEST
& BIOT SBS
& 69.07 & 69.89 & -- \\

FOG
& TEST
& Competence only
& $69.55\pm0.13$
& $70.22\pm0.10$
& $+0.33$ \\

FOG
& TEST
& Transfer only
& $71.60\pm0.36$
& $69.90\pm0.48$
& $+0.01$ \\

FOG
& TEST
& Transfer aware
& $\mathbf{72.37\pm0.32}$
& $\mathbf{71.91\pm0.43}$
& $\mathbf{+2.02}$ \\
\hline

ISRUC
& VAL
& CBraMod SBS
& 79.889 & 65.919 & -- \\

ISRUC
& VAL
& Competence only
& $78.602\pm0.089$
& $61.580\pm0.053$
& $-4.339$ \\

ISRUC
& VAL
& Transfer only
& $81.179\pm0.159$
& $64.544\pm0.075$
& $-1.375$ \\

ISRUC
& VAL
& Transfer aware
& $78.627\pm0.204$
& $61.314\pm0.247$
& $-4.605$ \\
\bottomrule
\end{tabular}
\end{table}

On FOG, transfer-aware routing improves over the BIOT SBS by $2.02\pm0.43$ balanced-accuracy points. Transfer-only routing rescues $1035.8\pm38.9$ BIOT errors but harms $826.6\pm47.6$ BIOT successes. Combining competence and transfer risk reduces the number of harmful switches to $307.4\pm80.1$ while retaining $580.4\pm69.9$ rescues, producing a net gain of $273.0\pm26.5$ correct predictions.

The result does not transfer to ISRUC. None of the three label-free variants exceeds the validation-selected CBraMod SBS. Transfer-only is the strongest label-free variant at $64.54\pm0.07\%$ balanced accuracy, but remains 1.38 points below CBraMod. The prespecified transfer-aware combination reaches only $61.31\pm0.25\%$.

These results suggest that transfer-derived signals can be informative within an individual pathway without being calibrated across heterogeneous foundation models. Native-teacher confidence, teacher-bridge divergence, and prediction disagreement may have different numerical meanings for different receivers. Consequently, the same fixed transfer-aware score does not provide a universally reliable estimate of downstream pathway utility.

Because the ISRUC label-free formulation did not satisfy its frozen validation criterion, it was not promoted to final test evaluation. The ISRUC values are therefore reported on VAL, whereas the previously frozen FOG experiment is reported on TEST. This experiment is included as an exploratory extension and negative cross-dataset result rather than as the primary method.

\subsection{FOG Direct-EEG Performance}
\label{app:fog-direct-eeg}

Table~\ref{tab:fog-direct-eeg} compares each frozen EEG model evaluated on
the paired EEG signal with the corresponding NormWear-to-EEG bridge evaluated
on EMG.  Direct-EEG performance does not determine cross-modal bridge
performance: CSBrain and BENDR improve after bridging, whereas LaBraM, EEGPT,
and REVE undergo substantial degradation and produce strongly class-biased
predictions.  This ranking mismatch motivates treating a bridge pathway as the
unit of selection rather than selecting a receiver solely from its native-EEG
performance.
\begin{table*}[htb!]
\centering
\caption{FOG direct-EEG and bridged-EMG balanced accuracy on the final test
split.  Values are percentages.}
\label{tab:fog-direct-eeg}
\small
\begin{tabular}{lrrr}
\toprule
Receiver & Direct EEG & Bridged EMG & Bridge $-$ direct \\
\midrule
LaBraM   & 66.02 & 49.98 & $-16.04$ \\
EEGPT    & 72.89 & 51.09 & $-21.80$ \\
REVE     & 72.59 & 50.41 & $-22.18$ \\
BIOT     & 76.89 & 69.89 & $-7.00$  \\
CBraMod  & 60.54 & 58.02 & $-2.52$  \\
CSBrain  & 58.25 & 64.05 & $+5.80$  \\
BENDR    & 58.50 & 65.96 & $+7.46$  \\
\bottomrule
\end{tabular}
\end{table*}

\subsection{FOG Output-Level Transfer Loss and Representation Distance}
\label{app:fog-transfer-loss}

We compare, for every paired instance and receiver, the class distribution
$p^{e}_{ik}$ produced by direct EEG with the class distribution $p^{b}_{ik}$
produced after bridge injection.  We report Jensen--Shannon (JS) divergence,
prediction-flip rate, and the AUC obtained when JS divergence is used to detect
a bridge failure conditional on a correct EEG teacher.  As shown in
Table~\ref{tab:fog-output-transfer}, output discrepancy is highly
receiver-dependent.  JS divergence is especially diagnostic for EEGPT and
REVE, but substantially less discriminative for CSBrain and BENDR.  A single
uncalibrated discrepancy threshold is therefore unlikely to transfer across
heterogeneous receivers.
\begin{table*}[htb!]
\centering
\caption{FOG output-level transfer diagnostics on the final test split.}
\label{tab:fog-output-transfer}
\begin{tabular}{l | rrrrr}
\toprule
Receiver & EEG BAcc & Bridge BAcc & Mean JS & Flip rate & JS AUC$_{\rm fail}$ \\
\midrule
LaBraM  & 66.02 & 49.98 & 0.0719 & 37.32\% & 0.9449 \\
EEGPT   & 72.89 & 51.09 & 0.1190 & 44.26\% & 0.9665 \\
REVE    & 72.59 & 50.41 & 0.1287 & 57.48\% & 0.9904 \\
BIOT    & 76.89 & 69.89 & 0.0554 & 32.39\% & 0.8260 \\
CBraMod & 60.54 & 58.02 & 0.1759 & 48.65\% & 0.8311 \\
CSBrain & 58.25 & 64.05 & 0.0149 & 38.16\% & 0.6365 \\
BENDR   & 58.50 & 65.96 & 0.0025 & 40.87\% & 0.6755 \\
\bottomrule
\end{tabular}
\end{table*}

For BIOT, we additionally compare the bridged representation with the paired
direct-EEG representation at bridge destination layer 2 and after propagation
through the remaining receiver layers.  Table~\ref{tab:biot-rep-distance}
shows nearly identical destination-layer discrepancy across the paired,
validation, and test splits.  The remaining BIOT layers contract rather than
amplify this discrepancy: test cosine distance decreases from 0.226 at the
destination to 0.134 at the final representation.  Nevertheless, a nontrivial
output discrepancy remains, with mean test JS divergence 0.055.
\begin{table*}[htb!]
\centering
\caption{BIOT representation-distance diagnostics.  ``Pair'' is the bridge
training domain; validation and test are held out from bridge fitting.}
\label{tab:biot-rep-distance}
\begin{tabular}{l | rrrrr}
\toprule
Split & Dest. NMSE & Dest. cosine & Final NMSE & Final cosine & Output JS \\
\midrule
Pair & 0.4915 & 0.2249 & 0.2428 & 0.1177 & 0.0520 \\
Validation & 0.4947 & 0.2263 & 0.2768 & 0.1341 & 0.0565 \\
Test & 0.4969 & 0.2262 & 0.2766 & 0.1341 & 0.0554 \\
\bottomrule
\end{tabular}
\end{table*}

\subsection{Collapsed Routes and Compact-Portfolio Headroom}
\label{app:fog-compact}

We define a route as collapsed using validation data only when one predicted
class accounts for at least 95\% of its outputs and validation balanced
accuracy is below 55\%.  This criterion removes LaBraM, EEGPT, and REVE and
retains BIOT, CBraMod, CSBrain, and BENDR.  
Table~\ref{tab:fog-collapse}
reports the validation-defined decision together with test behavior.
\begin{table*}[htb!]
\centering
\caption{Validation-defined FOG route-collapse audit.}
\label{tab:fog-collapse}
\begin{tabular}{l | rrcrr}
\toprule
Receiver & Val. dominant & Val. BAcc & Collapsed & Test dominant & Test BAcc \\
\midrule
LaBraM  & 99.95\% & 50.06 & Yes & 99.95\% & 49.98 \\
EEGPT   & 97.63\% & 51.38 & Yes & 97.88\% & 51.09 \\
REVE    & 99.03\% & 50.59 & Yes & 98.98\% & 50.41 \\
BIOT    & 52.79\% & 70.56 & No  & 52.30\% & 69.89 \\
CBraMod & 53.69\% & 58.67 & No  & 53.48\% & 58.02 \\
CSBrain & 59.11\% & 63.24 & No  & 58.76\% & 64.05 \\
BENDR   & 50.67\% & 64.75 & No  & 50.85\% & 65.96 \\
\bottomrule
\end{tabular}
\end{table*}

The full seven-route oracle reaches 99.96\% test balanced accuracy and misses
only three instances.  Removing the three collapsed routes reduces the oracle
to 93.69\% and increases the number of jointly missed instances to 514.
Thus, collapsed routes inflate the full oracle, but they do not explain away
portfolio complementarity: the compact-four oracle remains 23.80 points above
the BIOT SBS.  When the target-only MLP is added as the default predictor, the
compact-four residual oracle reaches $97.07\pm0.08$\% balanced accuracy across
the five MLP seeds, indicating that the retained bridges can correct 82--83\%
of the target-only MLP errors.

\subsection{FOG Supervised Controls and Higher-Capacity Routing}
\label{app:fog-controls}

Table~\ref{tab:fog-controls} compares the proposed gain-aware router with
target-only classifiers and a higher-capacity strict EEG-AS adaptation.  All
stochastic results use seeds 42--46 and report sample standard deviation.  The
target-only MLP is the strongest deployable method in this diagnostic.  This
result shows that the frozen target representation itself contains substantial
task information and limits any claim that bridge routing is universally
necessary.  BridgeRouter instead establishes the narrower result that
anchor-relative pathway supervision consistently improves the predefined SBS
while retaining single-route bridge execution.
\begin{table}[htb!]
\centering
\caption{FOG test controls.  ``Routes'' counts executed bridge pathways and
does not include the frozen target encoder.}
\label{tab:fog-controls}
\begin{tabular}{l | rrr}
\toprule
Method & BAcc & $\Delta$ vs. BIOT & Routes \\
\midrule
BIOT SBS & 69.89 & -- & 1.00 \\
Target-only linear & $79.74\pm0.17$ & $+9.85$ & 0 \\
BridgeRouter & $81.38\pm1.11$ & $+11.49$ & 1.00 \\
Strict EEG-AS adaptation & $82.19\pm0.47$ & $+12.30$ & 1.75 \\
Target-only MLP & $83.40\pm0.34$ & $+13.51$ & 0 \\
K7 oracle & 99.96 & $+30.07$ & 7.00 \\
\bottomrule
\end{tabular}
\end{table}

Strict EEG-AS jointly predicts missing route tokens and ranks the seven
pathways.  It improves over the simpler gain router by 0.81 balanced-accuracy
points on average, but it also increases mean route execution from 1.00 to
1.75.  Because its corresponding ISRUC evaluation was not completed under the
corrected CBraMod-SBS protocol, we treat it as a FOG-only higher-capacity
diagnostic rather than the primary cross-dataset method.

\subsection{ISRUC Development-Stage Sensitivity Analyses}
\label{app:isruc-development-sensitivities}

To assess sensitivity to the routing target, portfolio size, and source
representation, we report additional TRAIN-to-VAL analyses from the
development stage using a fixed anchor and seeds 42--46. Thresholds and
checkpoints are selected on VAL, and TEST is not accessed in any of these
analyses.

\paragraph{Routing target and SBS protection.}
Table~\ref{tab:isruc-development-targets} compares controlled changes to the
training target, router capacity, and protection rule while holding the earlier
H4/K7 development configuration fixed. Within this configuration, the
NLL-gain MLP with validation-selected protection gives the highest mean VAL
balanced accuracy. Correctness-gain routing selects thresholds that suppress
switching, whereas forcing alternative routing with the same NLL-gain scores
slightly reduces performance relative to retaining the anchor prediction.
Overall, these analyses support the use of validation-selected protection and
show that gains depend on the routing objective rather than switching alone.
\begin{table*}[htb!]
\centering
\caption{ISRUC routing-target and protection sensitivity under the earlier
H4/K7 development configuration. Results are VAL percentages over seeds
42--46; TEST was not accessed. $\Delta$BAcc is measured relative to the fixed
anchor baseline (64.9376 VAL balanced accuracy).}
\label{tab:isruc-development-targets}
\begin{tabular}{l | rrr}
\toprule
Condition & Accuracy & BAcc & $\Delta$BAcc \\
\midrule
NLL gain, MLP, protected
    & $78.60\pm0.36$ & $\mathbf{65.55\pm0.28}$ & $+0.61$ \\
Correctness gain, MLP, protected
    & $81.56\pm0.00$ & $64.94\pm0.00$ & $+0.00$ \\
True-class-probability gain, MLP, protected
    & $81.29\pm0.18$ & $65.04\pm0.11$ & $+0.11$ \\
NLL gain, linear, protected
    & $78.74\pm1.52$ & $65.14\pm0.09$ & $+0.20$ \\
NLL gain, MLP, forced alternative
    & $77.46\pm0.68$ & $64.91\pm0.22$ & $-0.03$ \\
\bottomrule
\end{tabular}
\end{table*}

\paragraph{Portfolio size.}
The predecessor portfolio study used a nested sequence constructed without
TEST access. Starting from a fixed anchor route, each additional route was
selected by the largest TRAIN subset-oracle balanced accuracy under a fixed
tie rule. Table~\ref{tab:isruc-development-k} shows that oracle balanced
accuracy increases monotonically as routes are added, whereas deployable
router performance is non-monotonic. The strongest exploratory router result
occurs at K4, while further increases in oracle headroom do not translate
directly into higher deployable performance. These development-stage
comparisons are sensitivity analyses only; neither the K4 result nor the
exploratory nested ordering replaces the final canonical K7 portfolio.

\begin{table*}[htb!]
\centering
\caption{ISRUC portfolio-size sensitivity under the earlier development
configuration. Nested subsets were constructed using TRAIN only and evaluated
on VAL; TEST was not accessed. Router values are mean $\pm$ sample standard
deviation over seeds 42--46. Route identities are shown only to document the
deterministic nested construction.}
\label{tab:isruc-development-k}
\begin{tabular}{c | l | rrr}
\toprule
$K$ & Nested portfolio & Oracle BAcc & Router BAcc & $\Delta$BAcc \\
\midrule
1 & BENDR & 64.94 & 64.94 & $+0.00$ \\
2 & + EEGPT & 71.10 & $64.95\pm0.04$ & $+0.02$ \\
3 & + LaBraM & 72.72 & $64.94\pm0.00$ & $+0.00$ \\
4 & + CBraMod & 75.94 & $\mathbf{65.77\pm0.26}$ & $+0.83$ \\
5 & + CSBrain & 76.48 & $65.61\pm0.15$ & $+0.67$ \\
6 & + REVE & 77.72 & $65.61\pm0.32$ & $+0.67$ \\
7 & + BIOT & 78.17 & $65.55\pm0.28$ & $+0.61$ \\
\bottomrule
\end{tabular}
\end{table*}

\paragraph{Source representation.}
Table~\ref{tab:isruc-development-layers} evaluates the source representation
while holding the predecessor K7 configuration fixed. Performance increases
gradually from H0 to a broad H3--H5 plateau and then declines. The difference
between H4 and H5 is only 0.0491 balanced-accuracy points, smaller than the
observed seed dispersion. Concatenating all eight layers increases the source
representation dimension from 512 to 4096 without improving on the
single-layer plateau, indicating limited sensitivity to the exact layer choice
within this range.
\begin{table}[htb!]
\centering
\caption{ISRUC HuBERT-ECG source-representation sensitivity under the earlier
K7 development configuration. Results are VAL balanced accuracy over seeds
42--46; TEST was not accessed.}
\label{tab:isruc-development-layers}
\begin{tabular}{l | r | r}
\toprule
Representation & Dimension & BAcc \\
\midrule
H0 & 512 & $65.2908\pm0.0726$ \\
H1 & 512 & $65.3383\pm0.1490$ \\
H2 & 512 & $65.4294\pm0.0512$ \\
H3 & 512 & $65.5254\pm0.2802$ \\
H4 & 512 & $65.5515\pm0.2820$ \\
H5 & 512 & $\mathbf{65.6006\pm0.2484}$ \\
H6 & 512 & $65.3949\pm0.2973$ \\
H7 & 512 & $65.3177\pm0.2331$ \\
H0--H7 concatenated & 4096 & $65.5234\pm0.1001$ \\
\bottomrule
\end{tabular}
\end{table}

\subsection{Rescue-Harm Diagnostics}
\label{app:rescue-harm}

The FOG router recovers many more BIOT errors than it destroys BIOT successes
(Table~\ref{tab:fog-rescue-harm}).  ISRUC presents a less stable tradeoff.  For
seeds 42--45, the router switches conservatively and achieves rescue precision
between 75.6\% and 94.0\%.  Seed 46 switches much more aggressively and its
rescue precision falls to 42.2\%, explaining much of the variance in the
aggregate ISRUC result.

\begin{table}[t]
\centering
\caption{FOG per-seed rescue--harm counts relative to the BIOT SBS.}
\label{tab:fog-rescue-harm}
\begin{tabular}{crrrr}
\toprule
Seed & Rescued errors & Harmed successes & Net correct & Switch rate \\
\midrule
42 & 1430 & 385 & $+1045$ & 68.73\% \\
43 & 1314 & 357 & $+957$  & 66.58\% \\
44 & 1336 & 440 & $+896$  & 67.40\% \\
45 & 1511 & 384 & $+1127$ & 73.93\% \\
46 & 1497 & 387 & $+1110$ & 72.99\% \\
\bottomrule
\end{tabular}
\end{table}

\begin{table*}[htb!]
\centering
\caption{ISRUC switching diagnostics relative to the CBraMod SBS.}
\label{tab:isruc-switching}
\begin{tabular}{crrrr}
\toprule
Seed & False switch$\mid$SBS correct & Attempted rescue$\mid$SBS wrong & Rescue precision & Net correct \\
\midrule
42 & 2.42\% & 9.24\% & 91.64\% & $+86$ \\
43 & 3.63\% & 12.18\% & 85.74\% & $+154$ \\
44 & 4.05\% & 14.01\% & 75.56\% & $+139$ \\
45 & 1.69\% & 2.80\% & 93.97\% & $-113$ \\
46 & 47.29\% & 53.40\% & 42.19\% & $+274$ \\
\bottomrule
\end{tabular}
\end{table*}

\subsection{Router Optimization and Reproducibility Details}
\label{app:router-details}

\paragraph{Router architecture and optimization.}
The router input is the pooled NormWear block-6 representation on FOG
(768 dimensions) and the pooled HuBERT-ECG block-4 representation on ISRUC
(512 dimensions).  The MLP uses hidden widths 256 and 128, GELU activations,
dropout 0.2, and one output for each non-anchor route.  We optimize with AdamW
(learning rate $3\times10^{-4}$, weight decay $10^{-4}$), batch size 256, a
maximum of 100 epochs, gradient-norm clipping at 5, and patience 15.  All
reported stochastic results use seeds 42--46.

For each route, NLL gains are clipped to their training-set 1st and 99th
percentiles and standardized before smooth-$L_1$ regression.  Instances are
weighted inversely to training class frequency.  At each epoch, the protection
threshold is selected on validation data from zero, 101 quantiles of the
nonnegative maximum predicted gain, and a value immediately above the observed
maximum.  Balanced accuracy determines the checkpoint; when balanced accuracy
ties, the larger threshold is preferred, yielding the more conservative
policy.  No threshold or checkpoint is reselected on the test set.

\paragraph{Bridge insertion locations.}
Table~\ref{tab:bridge-layers} lists the NormWear-to-EEG layer pairs used for
FOG.  Each target encoder and EEG backbone is frozen while the bridge is fit on
paired data.  The same validation-selected bridge checkpoint is then used for
all routing experiments.

\begin{table}[htb!]
\centering
\caption{FOG bridge source and destination layers.}
\label{tab:bridge-layers}
\begin{tabular}{lcc}
\toprule
Receiver & NormWear source block & EEG destination layer \\
\midrule
LaBraM  & 4 & 1 \\
EEGPT   & 5 & 5 \\
REVE    & 7 & 21 \\
BIOT    & 6 & 2 \\
CBraMod & 4 & 6 \\
CSBrain & 9 & 6 \\
BENDR   & 9 & 6 \\
\bottomrule
\end{tabular}
\end{table}

\paragraph{Protocol safeguards.}
The SBS is selected by validation balanced accuracy and frozen before test
evaluation.  This gives BIOT on FOG and CBraMod on ISRUC.  In particular,
EEGPT and BENDR are not retrospectively promoted to the ISRUC SBS despite
their stronger test performance.  All route probabilities and target features
are aligned by persistent sample identifier, and train, validation, and test
sample identifiers are checked for overlap.  The ISRUC split is subject
disjoint; the FOG BioX split is sample-wise and contains the same 12
participants in all four subsets.  The portfolio oracle uses the ground-truth
label and is reported only as a non-deployable upper bound.

\subsection{Inference Requirements and Computational Cost}
\label{app:inference-cost}

Table~\ref{tab:inference-requirements} distinguishes information used to train the router from information required at deployment.  
The gain targets used to train BridgeRouter are task-label-derived, but the trained router requires neither task labels nor the paired EEG modality at inference.  
It also does not execute candidate routes to make the routing decision.  
Instead, the frozen target encoder is evaluated once, the router predicts all
anchor-relative gains from that representation, and exactly one selected bridge pathway is executed. 
The K7 oracle evaluates every route and uses the ground-truth label, and is therefore reported only as an upper bound.
\begin{table}[htb!]
\centering
\caption{Information and pathway execution required at inference.  The target encoder is shared by SBS and BridgeRouter and is not counted as a bridge pathway.}
\label{tab:inference-requirements}
\begin{tabular}{l | cccc}
\toprule
Method & Target signal & Paired EEG & Task label & Bridge routes \\
\midrule
Validation-selected SBS & Yes & No & No & 1 \\
BridgeRouter & Yes & No & No & 1 \\
K7 oracle & Yes & No & Yes & 7 \\
\bottomrule
\end{tabular}
\end{table}

For the seven-route portfolio, the router predicts six gains because the SBS
is represented implicitly by the fallback decision.  With hidden widths 256
and 128, the router contains approximately 0.231M parameters for the
768-dimensional FOG input and 0.165M parameters for the 512-dimensional ISRUC
input.  This overhead is small relative to the frozen encoders and EEG
foundation models.  Most importantly, adaptive routing does not multiply
bridge-pathway inference cost by the portfolio size.

\end{document}